%% file: main.tex
\documentclass[conference]{IEEEtran}
\IEEEoverridecommandlockouts
\usepackage{cite}
\usepackage{amsmath,amssymb,amsfonts}
\usepackage{algorithmic}
\usepackage{graphicx}
\usepackage{textcomp}
\usepackage[table]{xcolor}
\usepackage{hyperref}
\usepackage{tikz}
\usetikzlibrary{shapes.geometric, arrows.meta, positioning, fit, backgrounds}
\usepackage{stfloats}
\usepackage{booktabs}
\usepackage{tabularx}
\usepackage{multirow}
\usepackage{caption}
\usepackage{rotating}

\begin{document}

\title{FineREX: Fine-Tuned NER-RE for Human Smuggling Knowledge Graphs}

\author{
    \IEEEauthorblockN{Elijah Feldman\IEEEauthorrefmark{1}\IEEEauthorrefmark{2},
    Dipak Meher\IEEEauthorrefmark{2},
    Carlotta Domeniconi\IEEEauthorrefmark{2}}

    \IEEEauthorblockA{\IEEEauthorrefmark{1}Thomas Jefferson High School for Science and Technology, Annandale, VA, USA}

    \IEEEauthorblockA{\IEEEauthorrefmark{2}Department of Computer Science, George Mason University, Fairfax, VA, USA\\
    \{efeldma5, dmeher, cdomenic\}@gmu.edu}
} \maketitle

\begin{abstract}
Court proceedings contain valuable evidence about human smuggling networks, but this information is often buried within unstructured, jargon-heavy legal documents. While large language models (LLMs) can support knowledge graph construction through automated information extraction, existing approaches rely on general-purpose models that are not tailored to the entity and relationship definitions required in this domain. We introduce FineREX, a streamlined knowledge graph construction pipeline built around a fine-tuned LLM for named entity recognition and relationship extraction (NER-RE). Using a manually annotated dataset of $512$ text chunks, FineREX achieves absolute improvements of $15.50\%$ and $31.46\%$ in entity and relationship F1-score, respectively, compared to a larger general-purpose baseline. These gains translate into higher-quality knowledge graphs, reducing legal noise by nearly half and lowering node duplication on long documents from $17.78\%$ to $11.17\%$. By eliminating document rewriting and redundant extraction stages, FineREX also reduces end-to-end processing time by $50.0\%$. Our results demonstrate that domain-specific fine-tuning can substantially outperform larger general-purpose models while improving both the quality and efficiency of knowledge graph construction for illicit network analysis. Our code is available 
\href{https://github.com/ElijahFeldman7/FineREX}{here}.

\end{abstract}


\section{Introduction}
\label{sec:intro}
Human smuggling is facilitated by decentralized and evolving networks of actors, resources, and transportation routes. Valuable intelligence about these operations is contained within court proceedings and other investigative records. However, the information is often fragmented across multiple documents and scattered throughout lengthy narratives, making it difficult for analysts and law enforcement agencies to reconstruct the full network structure. As a result, identifying critical actors, uncovering hidden relationships, and effectively disrupting smuggling operations remain significant challenges. \cite{dandurand2020migrant}. 

Large language models (LLMs) have shown considerable promise in synthesizing information from complex sources, including legal proceedings \cite{sinha2025legal} and scientific literature \cite{asai2026synthesizing}. However, these models remain susceptible to hallucinations \cite{Farquhar2024,mallen-etal-2023-trust}. In long documents, information appearing in the middle sections is less likely to receive sufficient attention \cite{bito2025evaluating}, increasing the risk that important facts and relationships will be overlooked. When reasoning across multiple documents, LLMs often struggle to perform inference in the presence of large amounts of distracting or irrelevant information \cite{belem2025single}. Furthermore, repeatedly querying LLMs over the same collection of documents is computationally inefficient and offers limited transparency and controllability, for directly exploring and analyzing the underlying networks.

   

These challenges motivate the use of structured named entity recognition and relation extraction (NER-RE). Structured extraction approaches have been shown to improve information extraction quality \cite{dagdelen2024structured} and enable documents to be partitioned into smaller segments for independent processing, thereby reducing the likelihood that important information is missed. In addition, LLMs are highly effective at identifying and linking relevant entities through relation extraction \cite{efeoglu2024relation}, helping uncover connections among key individuals and organizations.

In the context of human smuggling investigations, representing the extracted information as a knowledge graph (KG) offers significant advantages for both visualization and quantitative analysis. Given the inherently interconnected nature of smuggling operations, entities can be represented as nodes and their relationships as edges within a graph structure. This representation facilitates graph-based exploration and retrieval-augmented generation (RAG) approaches for querying \cite{edge2024local}, while also enabling network analysis techniques such as community detection to identify clusters of closely related actors and activities.


Prior work on knowledge extraction for human smuggling investigations \cite{meher2025linkkg, meher2025corekg, meher2025corekg_framework} has demonstrated the effective construction of knowledge graphs through multi-stage extraction pipelines. These approaches rely on general-purpose language models guided by descriptive system prompts to identify entities and relationships from unstructured text. However, \cite{skylaki2021legal} show that fine-tuning language models can substantially improve NER-RE performance, particularly in the legal domain, where models must distinguish relevant information from large amounts of domain-specific terminology and jargon. Despite these findings, the potential benefits of model fine-tuning remain largely unexplored in the NER-RE stages of human smuggling knowledge graph construction. Improving these stages is critical, as the quality of entity and relation extraction directly affects the completeness and accuracy of the resulting knowledge graphs.



LINK-KG \cite{meher2025linkkg}, a recent approach in this line of work, introduces a coreference resolution mechanism to address the challenge of entities being referenced by multiple names or aliases throughout a collection of documents. This process first requires an initial NER-RE stage to identify entities and relationships and populate a cache of resolved references. The cache is then used to rewrite the original text, after which a second NER-RE pass is performed to construct the final knowledge graph using a GraphRAG framework.

While effective, this approach incurs a substantial computational cost. Rewriting entire documents with an LLM is expensive, and performing NER-RE again on the resolved text largely duplicates the extraction effort already carried out during the initial pass. As a result, a significant portion of the computational budget is spent reprocessing information that has already been identified.

In this paper, we introduce FineREX (Fine-tuned Relation and Entity eXtraction), a streamlined knowledge graph construction pipeline that eliminates the text rewriting stage and the subsequent NER-RE pass required by prior approaches. Instead, FineREX employs a consolidation mechanism that directly leverages the coreference cache to resolve entities and relationships extracted during the initial NER-RE stage. This design reduces computational overhead while preserving the benefits of coreference resolution.

To improve extraction quality, we fine-tune the LLaMA 3.1 8B model specifically for NER-RE in the human smuggling domain. Since no existing dataset provides the entity and relationship definitions required for this task, we manually annotated 512 text segments ranging from one to three sentences in length, yielding a dataset containing 1,962 entities and 2,591 relationships for model training and evaluation. Experimental results show that FineREX achieves substantially higher entity- and relationship-level F1 scores than the baseline LLaMA 3.3 70B model. Furthermore, across sixteen U.S. Department of Justice human smuggling cases, FineREX significantly reduces the proportion of extracted legal noise, resulting in cleaner and more informative knowledge graphs.

Our key contributions are as follows:
\begin{itemize}
    \item We curate and manually annotate a specialized sentence-level NER-RE dataset comprising 512 samples, following entity and relationship definitions tailored to the human smuggling domain (Sec.~\ref{sec:dataset}).

    \item We introduce \textbf{FineREX}, a streamlined extraction framework that eliminates the text-rewriting stage and redundant NER-RE pass (Sec. \ref{sec:methods}). To improve extraction quality, we perform parameter-efficient fine-tuning of LLaMA 3.1 8B using our annotated dataset (Sec. \ref{sec:ft}).

    \item We conduct extensive empirical evaluations demonstrating that FineREX significantly outperforms both the untuned base model and a larger general-purpose model on NER-RE tasks (Sec. \ref{sec:ner-reresults}). When applied to complete case documents, FineREX reduces legal noise by nearly 50\% and decreases end-to-end runtime by 50.0\% relative to the baseline framework (Sec. \ref{sec:case-levelresults}).
\end{itemize}

\section{Related Work}
\label{sec:rw}

\subsection{Named Entity and Relationship Extraction}
\label{sec:nerre}
Early information extraction approaches relied on rule-based parsing to transform 
unstructured chat logs and notes into structured JSON representations 
\cite{santanna2020generating, min2025towards}. These methods often employed conditional random fields and word embeddings to model textual patterns. While effective on well-structured inputs, they struggle to capture implicit relationships and often fail when language deviates from expected patterns.

LLMs provide a more flexible alternative and have demonstrated strong performance 
on structured extraction tasks. Recent frameworks have applied general-purpose LLMs with prompt engineering for NER-RE  
\cite{pozzi2025kg_nlp_im, meher2025corekg_framework}. However, because these models are not specifically optimized for the extraction task, they generally underperform compared to task-adapted models. Prior work has shown that fine-tuning can substantially improve extraction performance, particularly in the biomedical domain \cite{gao2024joint} 
and in linguistic analysis tasks \cite{hicke2024literary_coref}, while QLoRA enables parameter-efficient adaptation.

Some approaches further decompose extraction into multiple stages. For example, LLMLink \cite{zhu2025llmlink} employs separate LLMs for entity extraction and relationship extraction. The authors observe that this design can introduce 
hallucinations that propagate between stages, suggesting that reducing the number of LLM inference steps may improve reliability. This motivates joint extraction approaches that identify entities and relationships simultaneously, leveraging the model's contextual understanding of entities to infer their interactions more effectively.

Despite these advances, no fine-tuned extraction models have been developed 
specifically for human smuggling cases, where legal documents often contain nuanced 
language, domain-specific terminology, and complex entity interactions.


\subsection{Knowledge Graph Construction}
\label{sec:kg}Recent LLM-based approaches have demonstrated strong performance for extracting structured information from unstructured text \cite{carta2023iterative, zhou2026finetuned}. These methods typically process documents in chunks to mitigate attention dilution and context-window limitations. For example, Zhou et al. \cite{zhou2026finetuned} show that a compact open-source model can effectively extract relational triplets from clinical text. However, chunking introduces challenges, as relationships spanning chunk boundaries may be missed when pipelines do not maintain information across segments.

A key step in knowledge graph construction is coreference resolution, which merges multiple mentions of the same real-world entity (e.g., \texttt{Sai Deshpande}, \texttt{Mr. Deshpande}, and \texttt{Sai}). Without coreference resolution, entity references become fragmented across multiple graph nodes, reducing graph quality and interpretability. In the human smuggling domain, CORE-KG introduced LLM-based coreference resolution, while LINK-KG \cite{meher2025linkkg} improved consistency through a persistent entity-mapping cache. LINK-KG combines chunk-level NER-RE, iterative coreference mapping, document rewriting, and a second GraphRAG-based extraction pass to construct the final knowledge graph.

Although effective, these frameworks rely on general-purpose LLMs for extraction and require multiple inference stages. Prior work has shown that knowledge graph quality is highly dependent on extraction accuracy, while additional LLM stages can introduce hallucinations and computational overhead \cite{zhu2025llmlink}. FineREX addresses both limitations by introducing a fine-tuned NER-RE model for human smuggling cases and a graph consolidation procedure that applies coreference mappings directly to extracted entities and relationships, eliminating document rewriting and redundant extraction passes. Furthermore, knowledge graphs can themselves support more reliable downstream reasoning, with prior work showing that grounding LLMs using graph-structured knowledge can reduce hallucinations \cite{wen2024mindmap}.

\begin{figure*}[htbp]
    \centering
    \includegraphics[width=\textwidth]{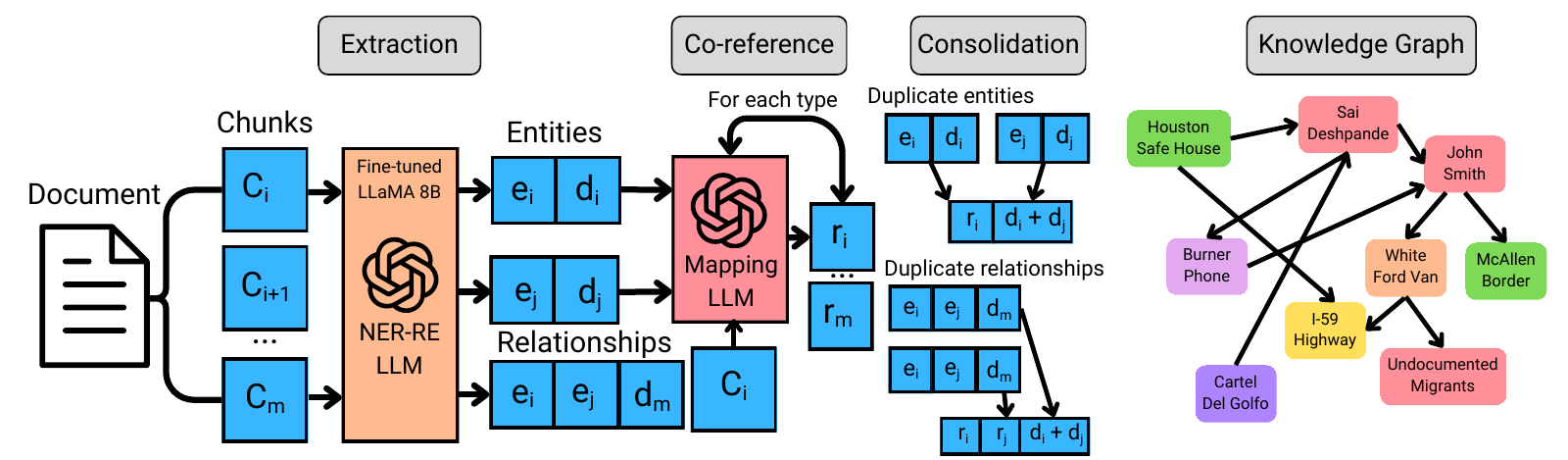}
    \caption{\textbf{Overview of FineREX}: Legal text is first chunked into segments of equal token length. The fine-tuned LLM extracts entities and relationships in a structured delimiter separated format. The extracted entities are provided to a coreference module that uses a Mapping-LLM to tie references to a unified name. Finally, we consolidate the NER-RE extractions by combining nodes that have the same canonical name and relationships that link two canonical entities.}
    \label{fig:pipeline}
\end{figure*}

\section{Methodology}
\label{sec:methods}
The proposed framework, FineREX, introduces a streamlined pipeline for constructing knowledge graphs from unstructured legal documents. Our approach relies on a single pass of a fine-tuned LLM followed by a LLM coreference process, and the subsequent consolidation of duplicate nodes. 

The overall architecture of our methodology is structured into four distinct, sequential phases as illustrated in Fig.\ref{fig:pipeline}:
\begin{enumerate}
    \item \textbf{Text Chunking:} Court documents are partitioned into segments of equal length to fit within the context window limits of the LLM and to prevent attention dilution over long spans of text.
    \item \textbf{NER-RE Extraction:} Each text chunk is processed by our fine-tuned LLaMA-3.1-8B-Instruct model. The model extracts entities and their directed relationships, outputting them in a structured delimiter separated format. 
    \item \textbf{Coreference Mapping:} The extracted entities are given to a coreference module created by Meher et al. \cite{meher2025linkkg} that uses a Mapping LLM to construct a reference dictionary. This process groups varied mentions under the same canonical name.
    \item \textbf{Graph Consolidation:} Using the coreference dictionary, duplicate nodes across different text chunks are merged. This step resolves conflicting properties such as canonical entities with different entity types and takes the mean relationship strength score to assemble the final knowledge graph
\end{enumerate}
The following subsections detail our dataset annotation, the fine-tuning procedure, and the processes of coreference resolutions and graph consolidation.

\subsection{Dataset}
\label{sec:dataset}
To the best of our knowledge, no prior dataset provides entity and relation extraction annotations for human smuggling cases. Furthermore, no existing dataset adopts the entity and relation schema required by our knowledge graph pipeline. This required the manual creation of ground-truth annotations, which was challenging due to the syntactically rigid nature of the text and the large number of entities and relations that could be identified within each chunk. Our dataset consisted of 1-3-sentence chunks, as this reasonably mimicked the amount that would be processed by the model during investigative analysis. The chunks were sampled from publicly available federal and state court documents retrieved with institutional access from Nexis Uni on human smuggling from 1994 to 2024. Due to Nexis Uni licensing restrictions, the raw dataset can not be redistributed. To avoid irrelevant procedural information, we only sampled from the Opinion section of these documents, which provide detailed descriptions of actors and relationships in each case. 

A team of two student annotators manually produced the ground-truth annotations for 512 samples. We categorize entities into several domain-relevant types, including \textit{Person}, \textit{Location}, \textit{Means of Transportation}, \textit{Means of Communication}, \textit{Route}, \textit{Organization}, and \textit{Smuggled Item}. Among these categories, \textit{Person} and \textit{Location} are the most prevalent, accounting for $45.6\%$ ($895$ entities) and $28.9\%$ ($567$ entities) of all annotated entities, respectively. This distribution reflects the prominence of individuals and locations in human smuggling court documents.

Given the prevalence of these entity types in the training data, the model is expected to perform particularly well at identifying \textit{Person} and \textit{Location} entities, as well as the relationships involving them. Overall, the dataset contains $1,962$ annotated entities connected by $2,591$ relationships.


The ground truth used the symbol  ``\texttt{|}" as a tuple delimiter to separate fields within each extracted entity or relationship, \texttt{\textbackslash n} to denote a record delimiter separating relationship or entity entries, and \texttt{<END>} to denote an end delimiter signaling the end of the generated output text. 

Entities are extracted in the format \texttt{("entity" | name | type | description)}, while relationships between two entities $e_i$ and $e_j$ are structured as \texttt{("relationship" | $e_i$ | $e_j$ | score | description)}.
The relationship score ranges from 0 to 10, with higher scores indicating a more explicitly defined relationship. Scores between 0 and 3 denote a weak relationship, which use uncertain phrasing or are implicit. Scores between 4 and 6 denote relationships that are explicitly stated however lack detailed context or additional information. Scores between 7 and 10 represent strong relationships that are clearly explained without hedging terms.

To prevent unnecessary downstream resolution, if an entity's reference includes a title such as "Driver Deshpande", the title "Driver" is provided in the description, while the model extracts "Deshpande" for the entity's name. 

To evaluate model performance under different data partitions, we use repeated random subsampling across thirty runs. For each run, the $512$ sample dataset is randomly partitioned into an $80\%$ training set, $10\%$ validation set, and $10\%$ testing set. This split strategy is necessary since each one-to-three-sentence chunk contains a mixture of multiple entity types and their corresponding directed relationships. We cannot stratify seven entity types over only fifty-one chunks without over-constraining the dataset. We addressed this limitation by using repeated random subsampling over thirty runs which creates a representative evaluation in the aggregate.

\subsection{Fine-tuning Process}
\label{sec:ft}
We utilized the open-source meta-llama/Llama-3.1-8B-Instruct as our base model. For parameter-efficient fine-tuning, we applied Quantized Low-Rank Adaptation (QLoRA). Instead of directly updating a $m \times n$ weight matrix $W$, QLoRA freezes the base model's weights in 4-bit precision and approximates an adapter matrix $\Delta W$ through a decomposition into two matrices $A$ and $B$, where $A$ has dimensions $m \times r$ and $B$ has dimensions $r \times n$. 

 For the NER-RE extraction process we used the LINK-KG detailed system prompt \cite{meher2025linkkg} to ensure consistency in comparing the two frameworks. This prompt included definitions in the human smuggling domain for our target entity types, along with three model samples and their accompanying ground truth. Furthermore, the prompt instructed the model to adhere to the delimiter format strictly and to ignore procedural and government-related legal entities.

The model was trained for four epochs using LoRA with rank $r = 16$, scaling parameter $\alpha = 32$, and a dropout rate of $0.05$ applied to all attention and feed-forward projection layers. These hyperparameter settings follow the empirical recommendations reported in the QLoRA framework of Dettmers et al.~\cite{dettmers2023qlora}. Training employed the paged AdamW 8-bit optimizer with a learning rate of $2 \times 10^{-5}$ and a cosine learning rate decay schedule. Gradient accumulation over four steps resulted in an effective batch size of four. To reduce memory consumption, we enabled both bfloat16 precision and gradient checkpointing. For each training run, the model checkpoint achieving the lowest validation loss was retained. 


\subsection{Coreference Resolution and Consolidation}
\label{sec:coref}


To construct the final knowledge graph, we pass the NER-RE output produced by the fine-tuned model to a coreference resolution module that identifies and resolves duplicate entity mentions. The coreference process leverages a Mapping LLM, specifically LLaMA 3.3 70B, which is prompted with the extracted entities and their associated descriptions from each document chunk. For each entity type, the module maintains a separate dictionary that maps entity aliases to canonical names and generates descriptive profiles for each canonical entity.

The consolidation stage uses these dictionaries together with the extracted entities and relationships to produce a resolved knowledge graph. Specifically, each entity identified by the fine-tuned model is normalized and queried against the appropriate dictionary. If a corresponding canonical entity is found, the entity mention is replaced with its canonical form; otherwise, the original entity name is retained.

Once canonical entities have been resolved, information distributed across multiple references and document chunks is aggregated into a single graph node. Whenever an entity is encountered in a new chunk, its type (e.g., \textit{Person} or \textit{Location}) is recorded and any newly extracted descriptions are appended to the entity's accumulated profile. This process consolidates evidence from multiple mentions, producing a more complete representation of each entity and reducing redundancy within the final knowledge graph.
The final type $\hat{t}(e)$ is selected 
by majority vote across all chunks:
    $\hat{t}(e) = \arg\max_{t \in \mathcal{T}} \; \text{count}(e, t)$,
where $\mathcal{T}$ is the set of entity types, and $\text{count}(e, t)$ is the number of times entity $e$ was assigned type $t$ across all chunks. This makes the 
consolidation robust to occasional type misclassification in individual chunks. 
Duplicate description sentences are removed and the remaining sentences are 
concatenated to form the final entity description.

Relationships are consolidated in a similar way. For each pair of canonical entities 
that share a relationship, we collect all relationship descriptions and strength scores 
across chunks. Duplicate descriptions are removed and the remaining unique descriptions are concatenated. The final 
relationship strength $\bar{s}(e_i, e_j)$ is computed as
    $\bar{s}(e_i, e_j) = \left\lfloor \frac{1}{|K|} \sum_{k \in K} s_k(e_i, e_j) \right\rfloor$
where $K$ is the set of chunks in which the relationship $(e_i, e_j)$ was extracted 
and $s_k(e_i, e_j)$ is the strength score assigned in chunk $k$.
\section{Experiments}
To validate the effectiveness of the fine-tuning approach, we evaluate the model on two levels. The first is a comparison of the LINK-KG pipeline's performance with the simplified pipeline using the fine-tuned NER-RE model. The second is a comparison of node duplication and legal noise across 16 cases of human smuggling to examine their impacts on downstream graph creation. 
All experiments were conducted on NVIDIA A100 GPUs with 80 GB of memory.
\input{sentence_results}
\subsection{NER-RE Evaluation}
We perform the fine-tuning process $30$ times, each with different random samples of the training, testing, and validation splits which are 80\%, 10\%, and 10\% of the dataset respectively. 

We used two baseline models for comparing NER-RE accuracy, the LLaMA 3.1 8B model before fine-tuning and LLaMA 3.3 70B which was used in LINK-KG for extraction \cite{meher2025linkkg}. We use the same few-shot prompt for both NER-RE models that is specified in the LINK-KG framework, consisting of the same domain-specific type definitions and delimiter instructions. No additional resolution steps were applied before evaluation based on the ground truth annotations.

\subsubsection{Entity metrics}
\label{sec:entitymetrics}
We evaluate entity extraction using standard Precision (P), Recall (R), and F1-score. An extracted entity is a true positive only if its name and type exactly match those of the ground truth. Entities with spelling errors are not counted as true positives, since they are detrimental to the extraction. If an entity is misspelled but the relationships are spelled correctly, then both the information from the entity and relationships associated with it are lost. Since descriptions could be correct but different from ground truth, we do not include them in the F1 score calculation. We calculated the metrics globally across entities (micro scores) and independently for each target entity type (macro scores).

\subsubsection{Relationship metrics}
\label{sec:relationmetrics}
We evaluate relationship extraction using P, R, and F1-score. A true positive represents a correctly identified link between two entities that was present in the ground truth. Relationships are directed with a source and target entity, so a relationship must be in the correct order to be considered a true positive. Additionally, we calculated the mean absolute error (MAE) score for the relationship strength. Formally, let $s_{ij} \in [0,10]$ be the ground truth strength score between entities $e_i$ and $e_j$, and let $\hat{s}\in[0,10]$ be the predicted score. The MAE is defined over the set of true positive relationships ($R_{\text{TP}}$) as:
$\text{MAE} = \frac{1}{|R_{\text{TP}}|} \sum_{(e_i, e_j) \in R_{\text{TP}}} |s_{ij} - \hat{s}_{ij}|
$
For each metric, we report the mean value across the $30$ independent runs.
\input{table.tex}

\subsection{Case-level Evaluation}
\label{sec:caseeval}
To evaluate the impact of FineREX on the quality of the constructed knowledge graphs, we generated graphs from 16 U.S. federal and state court cases spanning the period from 1994 to 2024. The dataset comprises seven shorter documents (2,500 words or fewer) and nine longer documents (more than 2,500 words). For consistency across all experiments, we use a chunk size of $300$ tokens.

We first assess the effect of fine-tuning on the quality of the resulting knowledge graphs. To this end, we run the complete LINK-KG pipeline, replacing its original NER-RE component with our fine-tuned model. The fine-tuned model used in these experiments was randomly selected from the thirty training runs. We refer to this variant as \textit{Fine-LINK-KG}. Both LINK-KG and Fine-LINK-KG use LLaMA 3.3 70B for GraphRAG generation.

Next, we evaluate the impact of the streamlined graph construction approach introduced by FineREX. Unlike LINK-KG, which performs text rewriting followed by a second GraphRAG-based extraction pass, FineREX employs a graph consolidation mechanism that directly resolves entities and relationships using the coreference mappings. This eliminates the need for document rewriting and redundant extraction steps.

Finally, to quantify the computational efficiency of each approach, we record the execution time of every pipeline stage and aggregate these measurements to compute the total runtime per case.

To measure the resulting graph quality, we use the same two metrics proposed in \cite{meher2025linkkg}:

\subsubsection{Duplicate Nodes}
Duplicate nodes corresponding to the same real-world entity reduce graph quality and complicate downstream analysis. We manually identified clusters $C_i$ of nodes referring to the same entity and computed the number of duplicate nodes as:
    $\text{Duplicates} = \sum_{C_i} (|C_i| - 1)$.
We report the percentage of duplicate nodes relative to the total number of graph nodes.


\subsubsection{Legal Noise}
Court case documents contain substantial legal jargon and many entities that are purely procedural in nature (e.g., judges, courts, and legal proceedings). We classify these as \emph{legal noise} because they do not contribute meaningful information about the underlying smuggling network and instead clutter the graph, hindering downstream analysis. Following \cite{meher2025linkkg}, we quantify legal noise as the percentage of graph nodes corresponding to such entities.

\subsection{Statistical Testing}
\label{sec:stats}
To assess statistical significance, we performed corrected paired t-tests over 30 resampled runs. Because the train/validation/test splits overlap substantially, the independence assumption of the standard paired t-test is violated. We therefore applied the variance correction of Nadeau and Bengio \cite{nadeau2003inference}. Using an 80/10/10 split, the correction factor is $\frac{1}{n}+\frac{0.10}{0.80}$ with $n=30$.

Normality was assessed using the Shapiro--Wilk test. All metrics satisfied the normality assumption ($p>0.05$) after applying a logit transformation to the bounded precision, recall, and F1 scores; prior to transformation, only entity precision showed marginal evidence of non-normality ($p=0.05$). To account for multiple comparisons across 14 hypotheses, we controlled the false discovery rate using the Benjamini--Yekutieli procedure \cite{benjamini2001fdr_dependency}.



\section{Results}
\label{sec:results}
\subsection{NER-RE Results}
\label{sec:ner-reresults}
Table \ref{tab:global_results} reports  precision, recall, and F1-score aggregated across all entity and relationship types, together with the MAE for relationship strength prediction. Compared to the baseline 8B model,  the fine-tuned 8B model achieves statistically significant improvements in entity F1 (16.86\%, $p < 0.001$) and relationship F1 (29.74\%, $p < 0.001$). Relative to the 70B baseline, the fine-tuned model, despite having nearly nine times fewer parameters, improves  entity F1 by 15.5\% ($p < 0.001$) and relationship F1 by 27.3\% ($p < 0.001$), corresponding to relative gains of 25.17\% and 134.21\%, respectively. These improvements are primarily driven by substantial increases in entity recall (25.47\%, $p < 0.001$) and relationship recall (41.62\%, $p < 0.001$). 
The results suggest that the general-purpose models fail to identify a considerable fraction of the target entities and relationships, whereas fine-tuning on human smuggling case data substantially improves extraction coverage.

In contrast, precision differences between the fine-tuned model and the 70B baseline are small and not statistically significant for either entities ($0.08\%$, $p=1.0$) or relationships ($0.23\%$, $p=1.0$). This is consistent with the observed recall gains, as increasing the number of predicted entities and relationships naturally introduces additional opportunities for false positives.  Furthermore, the fine-tuned model achieves a  statistically significant reduction in relationship strength MAE decreasing from $0.7855$ to $0.3450$ (a $56.08\%$ relative reduction, $p < 0.001$). This result indicates a substantial improvement in the model's ability to estimate relationship strength accurately.

Table \ref{tab:pertype_results} reports the precision, recall, and F1-score for each entity type. The fine-tuned 8B model outperformed the 70B baseline in recall and F1-score across all entity categories except \textit{Means of Communication}. This lower performance is likely attributable to the scarcity of this entity type in the training data, where it represents only 0.6\% of all annotated entities. In addition, the 70B baseline frequently hallucinated \texttt{DATE} and \texttt{FEES} entities, despite these categories not being requested in the extraction prompt, whereas the fine-tuned model adhered closely to the specified entity schema. Due to space constraints, we omit the per-type results for the baseline 8B model.


\subsection{Case-level Results}
\label{sec:case-levelresults}
We evaluate our proposed pipelines by first examining the impact of fine-tuning on graph quality to the baseline LINK-KG framework \cite{meher2025linkkg}, and then assessing the efficiency gains of our streamlined approach. The \emph{full pipeline} refers to the complete LINK-KG workflow, including text resolution and the second GraphRAG-based NER-RE pass. 
Tables \ref{tab:short_cases} and \ref{tab:long_cases} report graph quality metrics for LINK-KG, Fine-LINK-KG, and FineREX on short and long court cases, respectively.

Fine-tuning substantially improves the ability to filter irrelevant legal entities. Compared to LINK-KG, Fine-LINK-KG reduces legal noise from $12.28\%$ to $6.99\%$ on short documents and from $17.57\%$ to $7.50\%$ on long documents. The effect on node duplication is mixed for short cases, where LINK-KG achieves a slightly lower 
duplication rate ($10.61\%$ versus $15.79\%$), but Fine-LINK-KG performs considerably 
better on long cases, reducing duplication from $17.78\%$ to $9.16\%$. These results 
suggest that the fine-tuned model produces more coherent extractions, 
particularly for larger and more complex documents.

FineREX further improves graph quality on short documents, reducing legal 
noise and node duplication by $39.06\%$ and $11.91\%$, respectively, relative to 
Fine-LINK-KG. On long documents, FineREX exhibits a modest degradation, with 
legal noise and node duplication increasing by $19.26\%$ and $21.94\%$, respectively. 
Nevertheless, its graph quality remains comparable to that of Fine-LINK-KG while 
requiring substantially less computation.

\input{time_table}

Having established the benefits of fine-tuning, we next examine runtime performance. As shown in Table \ref{tab:runtime_performance}, replacing the 70B NER-RE model in LINK-KG with our fine-tuned 8B model reduces average processing time from $85.38$ to 
$74.99$ minutes, a $12.19\%$ reduction. By additionally eliminating text resolution and the second GraphRAG extraction stage, FineREX further reduces runtime to $42.66$ minutes. This corresponds to a $43.1\%$ reduction relative to Fine-LINK-KG and a $50.0\%$ reduction relative to the original LINK-KG pipeline.

Overall, FineREX provides a favorable trade-off between graph quality and 
computational efficiency. It achieves the lowest legal-noise rates on short 
documents while maintaining graph quality comparable to Fine-LINK-KG on longer cases, 
all while reducing end-to-end runtime by approximately one half.

\section{Conclusion}
\label{sec:conclusion}
In this work, we presented FineREX, a streamlined knowledge graph construction pipeline centered on a model fine-tuned for named entity and relationship extraction from human smuggling court proceedings. Our results show that the fine-tuned 8B model substantially outperforms a general-purpose LLaMA 3.3 70B baseline, achieving absolute improvements of 15.50\% and 31.46\% in entity and relationship F1-score, respectively. These gains translate into higher-quality knowledge graphs, reducing legal noise by nearly half and lowering node duplication on long documents from 17.78\% to 11.17\%.

Beyond improving extraction accuracy, FineREX simplifies the knowledge graph construction process by eliminating document rewriting and redundant extraction stages. This streamlined design reduces the computational cost of graph generation by approximately 50\% while maintaining strong graph quality. Together, these results demonstrate that domain-specific fine-tuning and pipeline simplification can significantly improve both the effectiveness and efficiency of knowledge graph construction in legal domains.

By transforming fragmented information from court proceedings into structured, analyzable knowledge graphs, FineREX provides investigators and law enforcement agencies with a practical tool for exploring human smuggling networks and identifying key actors, intermediaries, transportation routes, and organizational structures that may otherwise remain obscured across large collections of legal documents.

\section{Limitations}
\label{sec:limitations}
This work has some limitations. First, constructing a gold-standard dataset is inherently challenging, and because annotations were produced by two student annotators, some labels may reflect subjective interpretations of ambiguous entities or implicit relationships. Second, the scarcity of \textit{Means of Communication} and \textit{Smuggled Item} entities in the source documents results in limited training data for these categories and correspondingly weaker extraction performance. Finally, during graph consolidation, all relationships between the same pair of entities are merged into a single edge. While preserving multiple typed relationships could support richer analysis, doing so would require additional LLM inference and increase computational cost.
\bibliographystyle{IEEEtran}
\bibliography{references}

\end{document}

%% file: sentence_results.tex
\captionsetup[table]{justification=raggedright, singlelinecheck=false}
\begin{table*}[t]
\input{global_results}
\vspace{8pt}
\input{pertype_results}
\end{table*}

%% file: global_results.tex
\centering
\setlength{\tabcolsep}{4pt} 
\rowcolors{1}{gray!10}{white}
\begin{tabularx}{\textwidth}{X ccc @{\hspace{12pt}} cccc}
\hiderowcolors
\multirow{2}{*}{\textbf{Model ($n=30$)}} & \multicolumn{3}{c}{\textbf{Entity}} & \multicolumn{4}{c}{\textbf{Relationship}} \\
\cmidrule(lr){2-4} \cmidrule(lr){5-8}
 & P & R & F1 & P & R & F1 & MAE \\
\midrule
\showrowcolors
Baseline 8B & 0.6640 {\scriptsize $\pm$0.0344} & 0.5515 {\scriptsize $\pm$0.0319} & 0.6022 {\scriptsize $\pm$0.0289} & 0.4336 {\scriptsize $\pm$0.0758} & 0.1780 {\scriptsize $\pm$0.0280} & 0.2516 {\scriptsize $\pm$0.0384} & 1.7102 {\scriptsize $\pm$0.2520} \\
Baseline 70B & 0.7752 {\scriptsize $\pm$0.0318} & 0.5113 {\scriptsize $\pm$0.0303} & 0.6158 {\scriptsize $\pm$0.0283} & 0.5320 {\scriptsize $\pm$0.0846} & 0.1512 {\scriptsize $\pm$0.0356} & 0.2344 {\scriptsize $\pm$0.0486} & 0.7855 {\scriptsize $\pm$0.0869} \\
Fine-tuned 8B & \textbf{0.7760}$^*$ {\scriptsize $\pm$0.0307} & \textbf{0.7660}$^{*\dagger}$ {\scriptsize $\pm$0.0338} & \textbf{0.7708}$^{*\dagger}$ {\scriptsize $\pm$0.0295} & \textbf{0.5343}$^*$ {\scriptsize $\pm$0.0482} & \textbf{0.5674}$^{*\dagger}$ {\scriptsize $\pm$0.0461} & \textbf{0.5490}$^{*\dagger}$ {\scriptsize $\pm$0.0381} & \textbf{0.3450}$^{*\dagger}$ {\scriptsize $\pm$0.1279} \\
\hiderowcolors
\multicolumn{8}{l}{\vspace{-2mm}} \\ 
\multicolumn{8}{p{\dimexpr\textwidth-2\tabcolsep}}{\footnotesize $^*$ Statistically significant improvement over the Baseline 8B (adjusted $p < 0.05$).} \\
\multicolumn{8}{p{\dimexpr\textwidth-2\tabcolsep}}{\footnotesize $^\dagger$ Statistically significant improvement over the Baseline 70B (adjusted $p < 0.05$).} \\
\end{tabularx}
\caption{\textbf{Overall Extraction Performance:} We compare the global sentence-level entity and relationship extraction performance with Precision (P), Recall (R), F1-score, and Mean Absolute Error (MAE) for relationship strength scoring. Metrics report the mean $\pm$ standard deviation calculated over $n=30$ evaluation iterations.}
\label{tab:global_results}

%% file: pertype_results.tex
\centering
\setlength{\tabcolsep}{4pt} 
\rowcolors{1}{gray!10}{white}
\begin{tabularx}{\textwidth}{X ccc @{\hspace{12pt}} ccc}
\hiderowcolors
\multirow{2}{*}{\textbf{Entity Type}} & \multicolumn{3}{c}{\textbf{Baseline (LLaMA 70B, $n=30$)}} & \multicolumn{3}{c}{\textbf{Fine-tuned (LLaMA 8B, $n=30$)}} \\
\cmidrule(lr){2-4} \cmidrule(lr){5-7}
 & P & R & F1 & P & R & F1 \\
\midrule
\showrowcolors
Person                  & 0.7513 {\scriptsize $\pm$0.0478} & 0.5099 {\scriptsize $\pm$0.0435} & 0.6068 {\scriptsize $\pm$0.0419} & \textbf{0.7726} {\scriptsize $\pm$0.0492} & \textbf{0.7953} {\scriptsize $\pm$0.0474} & \textbf{0.7833} {\scriptsize $\pm$0.0444} \\
Location                & 0.8074 {\scriptsize $\pm$0.0712} & 0.5251 {\scriptsize $\pm$0.0636} & 0.6335 {\scriptsize $\pm$0.0571} & \textbf{0.8149} {\scriptsize $\pm$0.0473} & \textbf{0.7515} {\scriptsize $\pm$0.0661} & \textbf{0.7803} {\scriptsize $\pm$0.0473} \\
Organization            & 0.7433 {\scriptsize $\pm$0.3431} & 0.2060 {\scriptsize $\pm$0.1456} & 0.3061 {\scriptsize $\pm$0.1865} & \textbf{0.7450} {\scriptsize $\pm$0.1194} & \textbf{0.6044} {\scriptsize $\pm$0.1749} & \textbf{0.6529} {\scriptsize $\pm$0.1233} \\
Means of Transportation & \textbf{0.8191} {\scriptsize $\pm$0.0882} & 0.5477 {\scriptsize $\pm$0.0756} & 0.6536 {\scriptsize $\pm$0.0713} & 0.7880 {\scriptsize $\pm$0.0792} & \textbf{0.8143} {\scriptsize $\pm$0.0871} & \textbf{0.7999} {\scriptsize $\pm$0.0783} \\
Routes                  & 0.7952 {\scriptsize $\pm$0.2006} & 0.8544 {\scriptsize $\pm$0.2038} & 0.8176 {\scriptsize $\pm$0.1904} & \textbf{0.8228} {\scriptsize $\pm$0.1389} & \textbf{0.9015} {\scriptsize $\pm$0.1235} & \textbf{0.8514} {\scriptsize $\pm$0.1063} \\
Smuggled Items          & \textbf{0.2861} {\scriptsize $\pm$0.4057} & 0.1761 {\scriptsize $\pm$0.2611} & 0.2074 {\scriptsize $\pm$0.2930} & 0.2783 {\scriptsize $\pm$0.3215} & \textbf{0.3539} {\scriptsize $\pm$0.3908} & \textbf{0.2919} {\scriptsize $\pm$0.3187} \\
Means of Communication  & \textbf{0.4889} {\scriptsize $\pm$0.4484} & \textbf{0.5111} {\scriptsize $\pm$0.4568} & \textbf{0.4967} {\scriptsize $\pm$0.4489} & 0.2917 {\scriptsize $\pm$0.3968} & 0.2556 {\scriptsize $\pm$0.3602} & 0.2463 {\scriptsize $\pm$0.3182} \\
\end{tabularx}
\caption{\textbf{Per-type Extraction Performance:} We compare the entity extraction Precision (P), Recall (R), and F1-score metrics for the baseline 70B and our fine-tuned model. For each entity-type, the mean $\pm$ standard deviation are calculated over $n=30$ independent evaluation iterations. }
\label{tab:pertype_results}

%% file: table.tex
\begin{table*}[htbp]
    \centering
    \resizebox{\textwidth}{!}{
    \begin{tabular}{l | ccc | ccc | ccc || ccc | ccc | ccc}
        \multirow{3}{*}{\textbf{Case}} & \multicolumn{9}{c||}{\textbf{Node Duplication}} & \multicolumn{9}{c}{\textbf{Noisy Nodes}} \\
        \cmidrule(lr){2-10} \cmidrule(l){11-19}
        & \multicolumn{3}{c|}{\textbf{LINK-KG}} & \multicolumn{3}{c|}{\textbf{Fine-LINK-KG}} & \multicolumn{3}{c||}{\textbf{FineREX}} & \multicolumn{3}{c|}{\textbf{LINK-KG}} & \multicolumn{3}{c|}{\textbf{Fine-LINK-KG}} & \multicolumn{3}{c}{\textbf{FineREX}} \\
        & Total & Dup & \% & Total & Dup & \% & Total & Dup & \% & Total & Noise & \% & Total & Noise & \% & Total & Noise & \% \\
        \midrule
        Case 01 & 46 & 7 & 15.22 & 58 & 9 & 15.52 & 55 & 8 & \textbf{14.55} & 46 & 3 & 6.52 & 58 & 2 & \textbf{3.45} & 55 & 3 & 5.45 \\
        Case 02 & 52 & 7 & 13.46 & 45 & 4 & \textbf{8.89} & 45 & 4 & \textbf{8.89} & 52 & 2 & \textbf{3.85} & 45 & 2 & 4.44 & 45 & 3 & 6.67 \\
        Case 03 & 22 & 3 & 13.64 & 17 & 4 & 23.53 & 25 & 2 & \textbf{8.00} & 22 & 6 & 27.27 & 17 & 3 & 17.65 & 25 & 0 & \textbf{0.00} \\
        Case 04 & 34 & 2 & \textbf{5.88} & 67 & 11 & 16.42 & 63 & 6 & 9.52 & 34 & 5 & 14.71 & 67 & 2 & 2.99 & 63 & 0 & \textbf{0.00} \\
        Case 05 & 21 & 3 & \textbf{14.29} & 31 & 8 & 25.81 & 30 & 5 & 16.67 & 21 & 4 & 19.05 & 31 & 2 & \textbf{6.45} & 30 & 3 & 10.00 \\
        Case 06 & 32 & 2 & \textbf{6.25} & 31 & 3 & 9.68 & 24 & 4 & 16.67 & 32 & 2 & 6.25 & 31 & 1 & 3.23 & 24 & 0 & \textbf{0.00} \\
        Case 07 & 36 & 2 & \textbf{5.56} & 28 & 3 & 10.71 & 26 & 6 & 23.08 & 36 & 3 & 8.33 & 28 & 3 & 10.71 & 26 & 2 & \textbf{7.69} \\
        \midrule
        \textbf{Average} & 34.71 & 3.71 & \textbf{10.61} & 39.57 & 6.00 & 15.79 & 38.29 & 5.00 & 13.91 & 34.71 & 3.57 & 12.28 & 39.57 & 2.14 & 6.99 & 38.29 & 1.57 & \textbf{4.26} \\
    \end{tabular}
    }
    \caption{\textbf{Graph Quality Performance on Short Cases:} Node duplication and legal noise for short cases for LINK-KG, Fine-LINK-KG, and FineREX}
    \label{tab:short_cases}

\end{table*}

\begin{table*}[htbp]
    \centering
    \resizebox{\textwidth}{!}{
    \begin{tabular}{l | ccc | ccc | ccc || ccc | ccc | ccc}
        \multirow{3}{*}{\textbf{Case}} & \multicolumn{9}{c||}{\textbf{Node Duplication}} & \multicolumn{9}{c}{\textbf{Noisy Nodes}} \\
        \cmidrule(lr){2-10} \cmidrule(l){11-19}
        & \multicolumn{3}{c|}{\textbf{LINK-KG}} & \multicolumn{3}{c|}{\textbf{Fine-LINK-KG}} & \multicolumn{3}{c||}{\textbf{FineREX}} & \multicolumn{3}{c|}{\textbf{LINK-KG}} & \multicolumn{3}{c|}{\textbf{Fine-LINK-KG}} & \multicolumn{3}{c}{\textbf{FineREX}} \\
        & Total & Dup & \% & Total & Dup & \% & Total & Dup & \% & Total & Noise & \% & Total & Noise & \% & Total & Noise & \% \\
        \midrule
        Case 08 & 40 & 5 & 12.50 & 28 & 4 & 14.29 & 33 & 2 & \textbf{6.06}  & 40 & 10 & 25.00 & 28 & 2 & \textbf{7.14} & 33 & 4 & 12.12 \\
        Case 09 & 80 & 15 & 18.75 & 61 & 4 & \textbf{6.56}  & 39 & 3 & 7.69  & 80 & 16 & 20.00 & 61 & 6 & \textbf{9.84} & 39 & 5 & 12.82 \\
        Case 10 & 50 & 10 & 20.00 & 53 & 8 & 15.09 & 46 & 2 & \textbf{4.35}  & 50 & 6  & 12.00 & 53 & 5 & 9.43          & 46 & 1 & \textbf{2.17} \\
        Case 11 & 64 & 12 & 18.75 & 43 & 6 & 13.95 & 43 & 3 & \textbf{6.98}  & 64 & 4  & 6.25  & 43 & 7 & 16.28         & 43 & 2 & \textbf{4.65} \\
        Case 12 & 45 & 8  & 17.78 & 52 & 3 & \textbf{5.77}  & 49 & 7 & 14.29 & 45 & 17 & 37.78 & 52 & 1 & 1.92          & 49 & 0 & \textbf{0.00} \\
        Case 13 & 63 & 9  & 14.29 & 49 & 1 & \textbf{2.04}  & 45 & 6 & 13.33 & 63 & 15 & 23.81 & 49 & 1 & \textbf{2.04} & 45 & 6 & 13.33 \\
        Case 14 & 62 & 8  & 12.90 & 33 & 2 & \textbf{6.06}  & 30 & 3 & 10.00 & 62 & 6  & 9.68  & 33 & 1 & \textbf{3.03} & 30 & 3 & 10.00 \\
        Case 15 & 46 & 8  & 17.39 & 37 & 6 & \textbf{16.22} & 28 & 5 & 17.86 & 46 & 6  & 13.04 & 37 & 3 & 8.11          & 28 & 0 & \textbf{0.00} \\
        Case 16 & 76 & 21 & 27.63 & 40 & 1 & \textbf{2.50}  & 35 & 7 & 20.00 & 76 & 8  & 10.53 & 40 & 4 & \textbf{10.00} & 35 & 9 & 25.71 \\
        \midrule
        \textbf{Average} & 58.44 & 10.67 & 17.78 & 44.00 & 3.89 & \textbf{9.16} & 38.67 & 4.22 & 11.17 & 58.44 & 9.78 & 17.57 & 44.00 & 3.33 & \textbf{7.53} & 38.67 & 3.33 & 8.98 \\
    \end{tabular}
    }
    \caption{\textbf{Graph Quality Performance on Long Cases:} Node duplication and legal noise for long cases for LINK-KG, Fine-LINK-KG, and FineREX.}
    \label{tab:long_cases}
\end{table*}

%% file: time_table.tex
\begin{table}[t]
    \centering
    \renewcommand{\arraystretch}{0.9} 
    \resizebox{0.85\columnwidth}{!}{ 
    \begin{tabular}{l | rrr}
        \textbf{Case} & \textbf{LINK-KG} & \textbf{Fine-LINK-KG} & \textbf{FineREX} \\
        \midrule
        Case 1 & 45.34 & 38.72 & 16.51 \\
        Case 2 & 55.43 & 46.06 & 20.89 \\
        Case 3 & 57.31 & 48.46 & 22.85 \\
        Case 4 & 439.39 & 383.09 & 230.47 \\
        Case 5 & 73.26 & 67.13 & 33.22 \\
        Case 6 & 72.37 & 55.94 & 35.11 \\
        Case 7 & 96.62 & 86.49 & 46.76 \\
        Case 8 & 26.82 & 23.10 & 8.87 \\
        Case 9 & 25.43 & 20.49 & 8.09 \\
        Case 10 & 114.39 & 109.22 & 63.36 \\
        Case 11 & 28.04 & 20.12 & 8.89 \\
        Case 12 & 29.48 & 24.96 & 10.21 \\
        Case 13 & 68.99 & 53.32 & 31.98 \\
        Case 14 & 22.28 & 16.83 & 7.05 \\
        Case 15 & 84.21 & 91.69 & 52.05 \\
        Case 16 & 126.76 & 114.28 & 86.18 \\ 
        \midrule
        \textbf{Average} & \textbf{85.38} & \textbf{74.99} & \textbf{42.66} \\        
    \end{tabular}
    }
    \caption{ Processing time in minutes for each case  for LINK-KG, Fine-LINK-KG, and FineREX.}
    \label{tab:runtime_performance}
\end{table}